\documentclass{bmvc2k}
\usepackage{bmvc2k_natbib}
\usepackage{amsfonts}
\usepackage{algorithm}
\usepackage{algpseudocode}
\usepackage{booktabs}
\usepackage{multirow}
\usepackage{colortbl}
\usepackage{amssymb}
\usepackage{graphicx} 

\title{LidaRefer: Context-aware Outdoor 3D Visual Grounding for Autonomous Driving}

\addauthor{Yeong-Seung Baek}{bys414@koreatech.ac.kr}{1}
\addauthor{Heung-Seon Oh}{ohhs@koreatech.ac.kr} {1}

\addinstitution{
 School of Computer Science and Engineering \\
 Korea University of Technology and Education (KOREATECH)
}

\runninghead{}{}

\begin{document}

\maketitle

\begin{abstract}
3D visual grounding (VG) aims to locate objects or regions within 3D scenes guided by natural language descriptions. While indoor 3D VG has advanced, outdoor 3D VG remains underexplored due to two challenges: (1) large-scale outdoor LiDAR scenes are dominated by background points and contain limited foreground information, making cross-modal alignment and contextual understanding more difficult; and (2) most outdoor datasets lack spatial annotations for referential non-target objects, which hinders explicit learning of referential context. To this end, we propose \textbf{LidaRefer}, a context-aware 3D VG framework for outdoor scenes. LidaRefer incorporates an object-centric feature selection strategy to focus on semantically relevant visual features while reducing computational overhead. Then, its transformer-based encoder-decoder architecture excels at establishing fine-grained cross-modal alignment between refined visual features and word-level text features, and capturing comprehensive global context. Additionally, we present \textbf{Di}scriminative-\textbf{S}upportive \textbf{Co}llaborative localization (\textbf{DiSCo}), a novel supervision strategy that explicitly models spatial relationships between target, contextual, and ambiguous objects for accurate target identification. To enable this without manual labeling, we introduce a pseudo-labeling approach that retrieves 3D localization labels for referential non-target objects. LidaRefer achieves state-of-the-art performance on Talk2Car-3D dataset under various evaluation settings.
\end{abstract}

%-------------------------------------------------------------------------
\section{Introduction}
\begin{figure}[t]
  \centering
  \includegraphics[width = 0.95\textwidth]{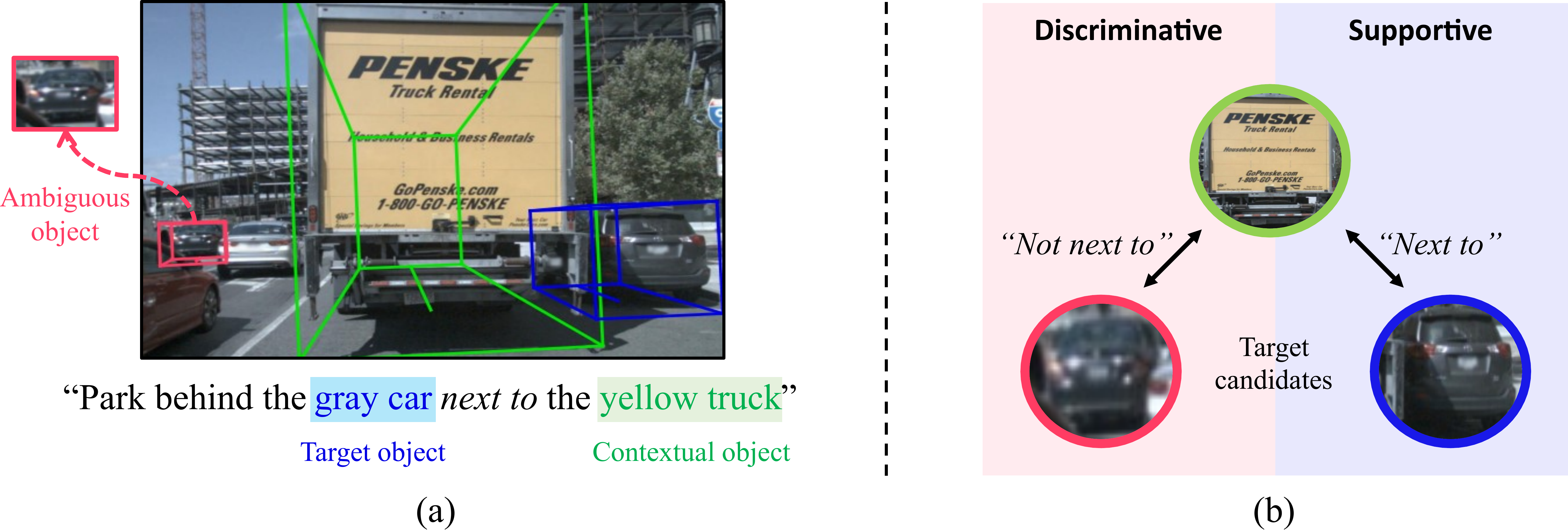}
  \caption{(a) Three types of key objects in VG scenarios. (b) Discriminative-Supportive Collaborative (DiSCo) approach focuses on two types of spatial relationships to effectively understand referential context: supportive relationships between target and contextual objects, and discriminative relationships between ambiguous and contextual objects.}
  \label{fig1}
\end{figure}

3D visual grounding (VG) aims to locate objects or regions within 3D visual scenes guided by natural language descriptions. This task is important for interactions between humans and agents in applications, such as autonomous driving, robotics, and VR/AR systems \cite{Sun2020, Park2020, Puig2018, Chen2020_1, Wijmans2019, Caesar2020}. While indoor 3D VG \cite{Huang2021, Yang2021, Yuan2021, Prabhudesai2020, Feng2021} has advanced, outdoor 3D VG with large-scale LiDAR point clouds remains underexplored despite its importance for autonomous driving, as in Fig.~\ref{fig1}(a).

In 3D VG scenarios, there are three types of key objects as in Fig.~\ref{fig1}(a): (1) target objects, which must be identified from a description; (2) contextual objects, which are described in the description and provide spatial references to the target; and (3) ambiguous objects, which resemble the target in appearance (e.g., category or even color), but are not the intended referent. Successful target identification requires two core capabilities. First, \textbf{cross-modal alignment}, which establishes semantic correspondence between textual and visual elements. Second, \textbf{contextual understanding}, which recognizes spatial relationships among objects within the 3D scene. For example, in Fig.~\ref{fig1}(a), with a description ("gray car next to the yellow truck"), to identify the target among similar candidates (red and blue boxes) in a scene, the model must align the textual references ("gray car" and "yellow truck") with their visual counterparts. Subsequently, it disambiguates the target (blue box) from ambiguous objects (red box) by recognizing spatial relationships ("next to") between target candidates and contextual objects (green box).

Recent methods leverage transformers \cite{Vaswani2017} to establish fine-grained cross-modal alignment while capturing scene-level global context through their attention mechanisms \cite{Huang2022, He2021, Zhao2021}. Some research \cite{Jain2022, Wu2023, yang2024exploiting, Bakr2023} focuses on modeling spatial relationships between target and contextual objects to explicitly learn object-level referential context specified in language descriptions. However, these methods are primarily designed for indoor scenes and face challenges when applied to outdoor scenes due to two key limitations. First, there exists a domain gap between indoor and outdoor scenes \cite{jin2024tod3cap, wang2023uni3detr}. In contrast to indoor scenes where foreground points constitute the majority of the point cloud, outdoor LiDAR scenes are dominated by background points, with foreground points appearing in limited regions. This extreme data distribution hinders both cross-modal alignment and contextual understanding by obscuring critical object features. Moreover, transformer architectures struggle with the computational and memory demands imposed by the high-dimensional visual representations derived from large-scale point clouds. Second, there exists a lack of comprehensive spatial annotations. Unlike indoor datasets \cite{Chen2020, Achlioptas2020} that provide 3D bounding boxes for both target and contextual objects, most outdoor datasets annotate only target objects. This absence significantly restricts the direct application of existing explicit context modeling techniques that utilize contextual object information, thereby hindering the learning of referential context through such approaches.

To this end, we propose \textbf{LidaRefer}, a context-aware 3D VG framework, for autonomous driving in large-scale outdoor scenes. Specifically, LidaRefer extracts object-centric features from high-dimensional and noisy visual features derived from LiDAR inputs to filter out irrelevant information. This leads to a reduction in computational complexity and an improvement in feature quality. Based on these refined visual features and word-level text features, LidaRefer establishes fine-grained cross-modal alignment while capturing global context through its transformer-based encoder-decoder structure.

Additionally, as a key component of LidaRefer, we introduce \textbf{Di}scriminative-\textbf{S}upportive \textbf{Co}llaborative localization, namely \textbf{DiSCo}, a novel supervision method that explicitly learns referential context. As in Fig.~\ref{fig1}(b), DiSCo focuses on two types of spatial relationships for referential context: (1) supportive relationships between target and contextual objects which provide essential spatial cues mentioned in the description, and (2) discriminative relationships between ambiguous and contextual objects, which reveal key differences in spatial positioning that enable the model to distinguish the target from visually similar candidates. By modeling these relationships, LidaRefer enables a precise understanding of referential context in complex scenarios, leading to accurate target identification. Since DiSCo requires 3D localization information for typically unlabeled contextual and ambiguous objects, and the manual labeling process is costly, we propose an efficient pseudo-labeling strategy. This strategy automatically provides the necessary 3D bounding boxes for these non-target objects, thereby enabling explicit referential context learning via DiSCo supervision without additional annotation burden.

Our contribution can be summarized as follows: 1) We propose LidaRefer, a context-aware 3D VG framework, suitable for large-scale outdoor scenes. 2) We introduce DiSCo, a novel supervision method that explicitly models and learns referential context, along with an efficient pseudo-labeling strategy. 3) We demonstrate LidaRefer's superiority by achieving state-of-the-art results on Talk2Car-3D dataset and analyzing the effects under various evaluation settings.

\section{Related Work}

\subsection{Outdoor 3D Visual Grounding}

Recent progress in 3D VG focuses on indoor scenes \cite{Guo2023, Huang2021, Huang2022, Yang2021, Yuan2021, Luo2022, Liu2021focal, Prabhudesai2020, Feng2021, He2021}, supported by various benchmark datasets \cite{Chen2020, Achlioptas2020, abdelreheem2024scanents3d, Zhang2023}. In contrast, outdoor 3D VG with large-scale point LiDAR point clouds is underexplored. Wildrefer \cite{lin2023wildrefer} proposes both a dataset and model utilizing LiDAR point clouds, but it focuses solely on humans and covers a relatively narrow range (i.e., 30m radius). Text2Pos \cite{kolmet2022text2pos} and CityRefer \cite{miyanishi2023cityrefer} introduce 3D VG datasets with city-scale LiDAR point clouds and their baselines. However, they have the limitation that a point cloud must be divided into smaller subsets for processing within a single scenario. Addressing this, MSSG \cite{Cheng2023} proposes a method suitable for autonomous driving environments by processing entire large-scale dynamic outdoor scenes at once. Yet, this approach relies on coarse-grained cross-modal alignment with sentence-level features, which may lose specific word meanings in complex descriptions, and offers limited consideration for contextual understanding. BEVGrounding \cite{liu2024talk}, also geared towards autonomous driving applications, adopts a transformer\cite{Vaswani2017}-based architecture to perform both coarse- and fine-grained cross-modal alignment. Nonetheless, it solely relies on target object information, restricting its ability to capture comprehensive referential context. Additionally, current outdoor 3D VG methods have not adequately addressed the inherent challenges of outdoor environments, where the prevalence of background points and empty spaces can destabilize learning processes, especially for both cross-modal alignment and contextual understanding.

\subsection{Context-aware Modeling}

Recognizing contextual information is crucial for accurate target identification in 3D VG scenarios. In indoor domains, several approaches have been developed to address this challenge. Early works \cite{Huang2021, Feng2021} employ Graph Neural Networks (GNNs) \cite{scarselli2008graph} to model object relationships as graphs, but are limited in capturing complex spatial dependencies. To overcome these limitations, several methods \cite{Huang2022, He2021, Zhao2021, Jain2022} use the transformer's attention mechanism. For instance, BUTD-DETR \cite{Jain2022} effectively captures global context while achieving fine-grained cross-modal alignment through its transformer-based encoder-decoder architecture. More recent approaches focusing on explicitly modeling referential context specified in language descriptions have gained attention. CORE-3DVG \cite{yang2024exploiting} introduces text-guided object detection and relation matching networks to extract both target and contextual objects while capturing their spatial relationships. BUTD-DETR \cite{Jain2022} and EDA \cite{Wu2023} incorporate localization supervision to identify all mentioned objects. CoT3DRef \cite{Bakr2023} reformulates 3D VG as a sequence-to-sequence task, leveraging the Chain-of-Thought \cite{wei2022chain} reasoning to predict a chain of contextual objects that guide target localization. Despite their benefits, these approaches face scalability challenges: they often depend on costly annotations or involve complex pseudo-labeling pipelines. This significantly limits their practicality in outdoor scenarios, where only target annotations are commonly available.

\section{Methodology}

\begin{figure}[t]
    \centering
    \includegraphics[width = 0.95\textwidth]{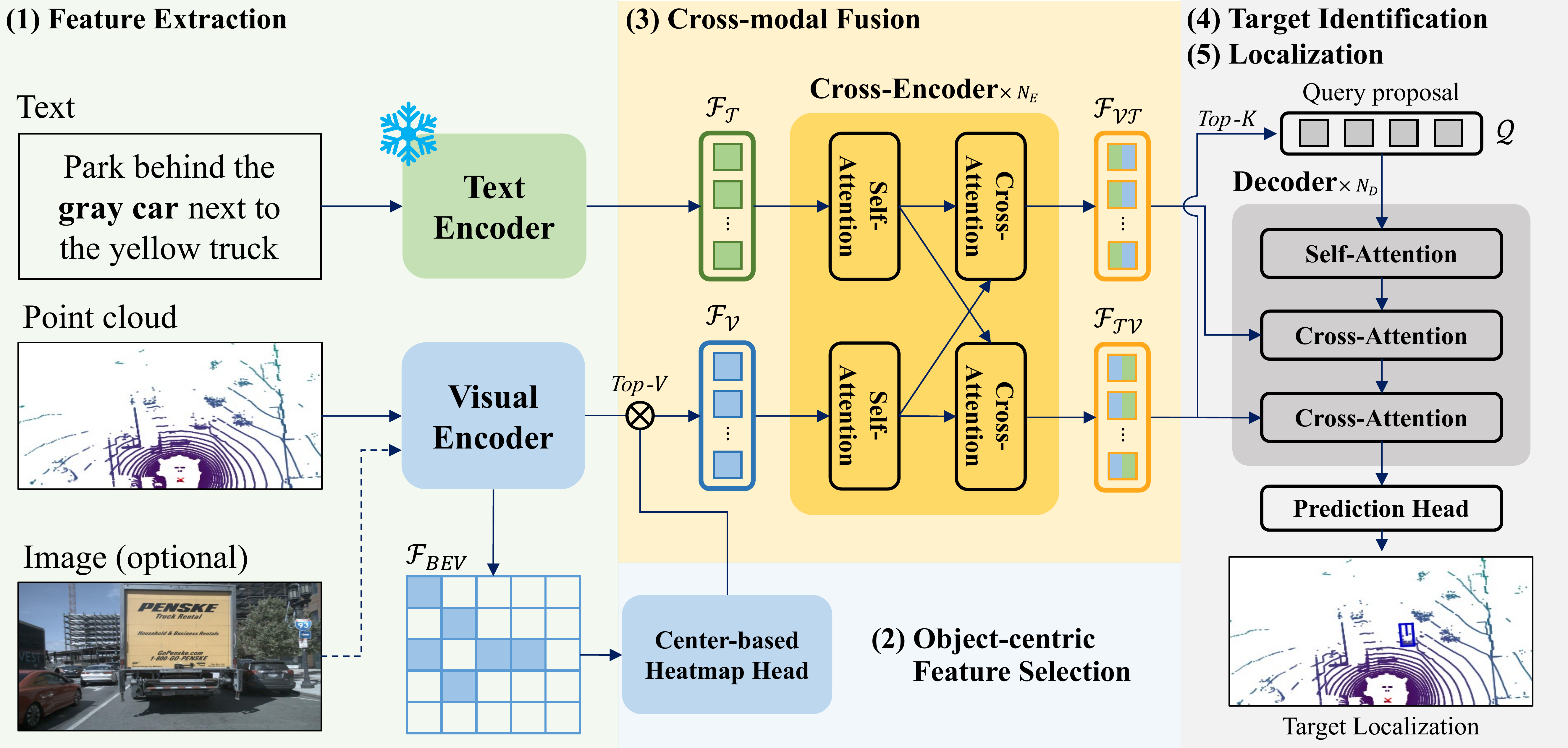} 
    \caption{The overview of LidaRefer. After extracting visual and text features, object-centric features are selected from high-dimensional visual features. Then, the features from both modalities are fed into a cross-modality encoder. Lastly, each query proposal feature is refined through a cross-modality decoder to predict a 3D bounding box for the target.}
    \label{fig2}
\end{figure}

Fig.~\ref{fig2} illustrates the overall architecture of LidaRefer, which takes synchronized inputs: a LiDAR point cloud and, optionally, RGB images as the visual input, along with a language description as the textual input. While our goal is the localization of the target object via a 3D bounding box based on the description, our training objective explicitly incorporates the localization of referential non-target objects (i.e., contextual or ambiguous ones) as an auxiliary task to alleviate ambiguity in target identification.

\subsection{Feature Extraction}

LidaRefer supports both LiDAR-only and multimodal (LiDAR + image) settings. In the LiDAR-only setting, points are encoded into 3D voxel features and projected onto the bird's-eye-view (BEV) plane using a standard voxel-based 3D backbone \cite{Yan2018}, chosen for its proven efficiency and effectiveness in various outdoor detectors \cite{Shi2020, Zhou2022, Yin2021}. In the multimodal setting, we adopt FocalsConv's \cite{Chen2022_focal} multimodal fusion strategy to effectively combine geometric features from point clouds with texture information from images. 2D image features are extracted using a ResNet-50 backbone \cite{He2016} pre-trained on MS-COCO \cite{Lin2014} detection task. These image features are fused into 3D voxel features through perspective transformation and summation operations. Then, the fused 3D voxel features are projected onto a BEV plane in the same manner as in the LiDAR-only setting. Let $\mathcal{F}_{\mathrm{BEV}} \in \mathbb{R}^{h \times w \times d}$, where $h$ and $w$ are the size of the BEV feature map and $d$ is the hidden size. 

A language description is encoded into word-level text features using pre-trained RoBERTa \cite{Liu2019}, employed in recent VG models \cite{Kamath2021,Jain2022,Wu2023,zhan2024mono3dvg} as a text encoder. This leverages prior knowledge about the words. Let $\mathcal{F}_{\mathcal{T}} \in \mathbb{R}^{l \times d}$ be the text features, where $l$ is the length of a description.

\subsection{Object-centric Feature Selection}

Outdoor LiDAR scenes are characterized by dominant background regions and sparse foreground objects distributed over a wide spatial area. Passing such high-dimensional and predominantly irrelevant visual features into transformer \cite{Vaswani2017} architectures leads to two critical issues: (1) excessive computational and memory overhead due to the quadratic complexity of attention, and (2) weakening the attention focusing on object-relevant regions, leading to unstable learning for cross-modal alignment and contextual understanding.

To address this, we employ an \textbf{O}bject-centric \textbf{F}eature \textbf{S}election strategy (\textbf{OFS}) inspired by outdoor 3D object detection approaches, effectively filtering out irrelevant background information. Following \cite{Zhou2022}, a center heatmap head over BEV features $\mathcal{F}_{\mathrm{BEV}}$ is applied to produce a category-wise heatmap $\mathcal{F}_{\mathrm{HM}} \in \mathbb{R}^{h \times w \times c}$, where each channel corresponds to one of $C$ categories. Then, we select the top-$V$ positions with the highest scores and extract their associated BEV features, yielding compact object-centric features $\mathcal{F}_{\mathcal{V}} \in \mathbb{R}^{v \times d} \subset \mathcal{F}_{\mathrm{BEV}}$. The heatmap head is trained using category and localization labels (i.e., center coordinates in BEV plane) for all objects in the scene, which can be readily obtained from off-the-shelf 3D object detectors or detection datasets commonly used as supersets of VG datasets. This selection mechanism reduces attention complexity while ensuring that transformers operate on spatially and semantically relevant visual inputs, improving both cross-modal alignment and contextual understanding in large-scale outdoor scenes.

\subsection{Cross-modal Fusion}

Once the object-centric visual features $\mathcal{F}_{\mathcal{V}}$ and the encoded text features $\mathcal{F}_{\mathcal{T}}$ are obtained, LidaRefer performs fine-grained cross-modal fusion using $N_E$ transformer-based cross-modality encoding layers to model both intra-modal and inter-modal interactions. Each encoding layer first adds modality-specific positional embeddings to visual and text tokens. Then, the visual and text tokens perform self-attention within each modality to capture global context and subsequently cross-attention with each other to achieve fine-grained cross-modal alignment. As a result, we obtain two aligned feature representations: text-aligned visual features $\mathcal{F}_\mathcal{TV} \in \mathbb{R}^{v \times d}$ and visual-aligned text features $\mathcal{F}_\mathcal{VT} \in \mathbb{R}^{l \times d}$. These aligned features serve as inputs to our transformer-based cross-modality decoder to predict the target object.

\subsection{Target Identification and Localization}

Similar to the query initialization in \cite{Jain2022, Wu2023}, the top-$K$ features are first selected from text-aligned visual features ($\mathcal{F}_{\mathcal{TV}}$) based on confidence scores computed by a 2-layer MLP as a query proposal network. These selected features are linearly projected to form query proposal features, which are subsequently refined via $N_D$ cross-modality decoding layers. Within each decoding layer, queries are enhanced with positional embeddings in the BEV plane, then self-attend to each other and sequentially cross-attend to $\mathcal{F}_{\mathcal{VT}}$ and $\mathcal{F}_{\mathcal{TV}}$. The final refined queries are passed through two task-specific heads: (1) a target identification head to predict a probability for each query of being the target object, and (2) a box regression head to predict a 3D bounding box for the target.

\subsection{Discriminative-Supportive Collaborative localization}

As in Fig.~\ref{fig1}(a), an ambiguity problem in 3D VG often arises when multiple objects share similar visual attributes with the target, making appearance-based clues insufficient for accurate identification. In such cases, understanding the referential context mentioned in descriptions is essential for resolving ambiguity. Although the proposed architecture implicitly captures this referential context during cross-modal fusion through its transformer-based cross-modal encoder, we argue that explicitly modeling spatial relationships more effectively leads to understanding the referential context. Specifically, we focus on two aspects of spatial relationships as in Fig.~\ref{fig1}(b): (1) supportive relationships between target and contextual objects, that align with the spatial configurations mentioned in the description (e.g., "gray car next to yellow truck"), and (2) discriminative relationships between ambiguous and contextual objects, which reveal how objects that visually resemble the target fail to satisfy the mentioned spatial configurations (e.g., positioning of other gray cars relative to the yellow truck). 

To this end, we introduce \textbf{Di}scriminative-\textbf{S}upportive \textbf{Co}llaborative localization (\textbf{DiSCo}), a novel training strategy to understand the referential context by simultaneously supervising the localization of target and referential non-target objects (i.e., contextual and ambiguous objects). DiSCo leverages query-level self-attention within the transformer-based decoder to facilitate spatial information exchange between key objects. Specifically, during training, decoder queries assigned to both target and referential non-target objects are encouraged to attend to each other, enabling the model to understand relative positions and spatial configurations among key objects within each scenario. This collaborative supervision approach enhances the model's comprehension of referential context, leading to accurate and robust target identification in challenging outdoor scenes with multiple similar objects.

To employ DiSCo, we require 3D bounding boxes for referential non-target objects as labels. However, such labels are typically unavailable in practical applications, and even manually labeling them can be prohibitively expensive. To address this, based on two observations, we introduce an efficient pseudo-labeling strategy to generate the pseudo-labels for referential non-target objects automatically. First, during query proposal, most queries are generated from target and referential non-target object locations since text-aligned visual features $\mathcal{F}_\mathcal{TV}$ in these regions exhibit high confidence due to their semantic correspondence to objects mentioned in the description. Second, the 3D bounding boxes for all objects in a scene can be readily retrieved using detection labels or off-the-shelf 3D detectors. Leveraging these signals, for each scenario, we assign 3D bounding boxes to referential non-target objects based on a center-based assignment approach typically used in outdoor 3D object detectors \cite{Zhou2022, Yin2021}. Specifically, we first match each decoder query $q_i \in Q = \{q_1, \dots, q_K\}$ to its nearest object proposal $b_j \in B = \{b_1, \dots, b_N\}$ by measuring the distance between their center coordinates, where $B$ represents the bounding boxes for all objects in the scene. Then, an object is assigned as a referential label if its distance to a query falls below a predefined threshold $\tau$. This lightweight assignment process enables DiSCo to supervise the referential context without incurring additional labeling costs. The overall DiSCo and the pseudo-labeling procedure are detailed in the supplementary material.

\section{Experiments}

\paragraph{Experiment Setting.} We briefly summarize our experimental setup\footnote{The details of our experiment setting are provided in the supplementary materials.}. We evaluate LidaRefer on Talk2Car-3D derived from Talk2Car \cite{Deruyttere2019}, a 2D VG dataset for autonomous driving built upon nuScenes \cite{Caesar2020}. Following \cite{Cheng2023, liu2024talk}, we adopt their preprocessing pipeline and data split methods to convert the original 2D settings to 3D for constructing Talk2Car-3D, where each scenario contains a synchronized LiDAR point cloud, an image, and a language description referring to a single object from one of 23 categories.

As an evaluation metric, we use Acc@$\mathrm{IoU}_{\mathrm{thr}}$, where a predicted box is considered correct if its 3D Intersection over Union (IoU) with the ground truth box exceeds a predefined threshold. Following \cite{liu2024talk}, we report the accuracies on 10 supercategories derived by merging semantically similar categories from the 23 categories under two threshold settings: 0.5 and 0.25. ACC@0.5 emphasizes precise localization more than ACC@0.25. Therefore, given similar performance in ACC@0.25, a higher ACC@0.5 score indicates superior localization ability. Conversely, when ACC@0.5 is comparable, a higher ACC@0.25 score reveals stronger identification performance.

We implement LidaRefer in two configurations based on input modalities: (1) a \textit{LiDAR-only setting} that only utilizes 3D point clouds to capture geometric structure and (2) a \textit{multimodal setting} that incorporates RGB images for texture information. In both settings, to assess the impact of prior knowledge of object-level semantics, we also evaluate a variant in which 3D visual encoders are initialized with weights pre-trained on nuScenes 3D object detection task.

We compare LidaRefer against recent state-of-the-art outdoor 3D VG baselines: MSSG \cite{Cheng2023} and BEVGrounding \cite{liu2024talk}. For BEVGrounding, we consider its two distinct variants: the baseline two-stage approach and a subsequently developed one-stage approach. Additionally, we include two random baselines: GT-Rand, which randomly selects a ground-truth annotated box in the scene as a prediction, and Pred-Rand, which randomly selects one of the predicted proposals as the output. Since MSSG does not report results under our evaluation settings, we report the LiDAR-only results through re-implementation\footnote{The original paper lacks the implementation details for the multimodal case.}. For BEVGrounding and the random baselines, we adopt the results reported in their paper.

\begin{table}[t]
    \begin{center}
        \scriptsize{%
            \begin{tabular}{lccccc}
                \toprule
                \multirow{2}{*}{Method} & Pre-trained & \multicolumn{2}{c}{BEV} & \multicolumn{2}{c}{3D} \\
                \cmidrule(lr){3-4} \cmidrule(lr){5-6} % BEV (3-4열), 3D (5-6열) 아래에 선 추가
                & 3D visual encoder & ACC@0.25 & ACC@0.5 & ACC@0.25 & ACC@0.5 \\
                \midrule
                Pred-Rand & - & 7.35 & 5.20 & 7.06 & 3.73 \\
                GT-Rand & - & 7.06 & 7.06 & 7.06 & 7.06 \\
                \midrule
                \multicolumn{6}{c}{\textit{LiDAR-only setting}} \\
                \midrule
                BEVGrounding-L (1-stage) \cite{liu2024talk}& - & 42.55 & 26.96 & 35.10 & 17.25 \\
                BEVGrounding-LP (2-stage) & \checkmark & 24.90 & 24.02 & 23.04 & 19.61 \\
                MSSG-L \cite{Cheng2023}& - & 43.67 & 28.31 & 41.97 & 24.10 \\
                MSSG-LP & \checkmark & 51.00 & \textbf{41.67} & 49.30 & \textbf{37.05} \\
                \rowcolor{gray!10} LidaRefer-L & - & 53.82 & 36.85 & 50.90 & 31.43 \\
                \rowcolor{gray!10} LidaRefer-LP & \checkmark & \textbf{56.83} & 40.96 & \textbf{54.42} & 36.85 \\
                \midrule
                \multicolumn{6}{c}{\textit{Multimodal setting}} \\
                \midrule
                BEVGrounding-M (1-stage) & - & 44.02 & 28.53 & 36.37 & 19.61 \\
                BEVGrounding-MP (2-stage) & \checkmark & 28.73 & 25.49 & 26.37 & 24.51 \\
                \rowcolor{gray!10} LidaRefer-M & - & 56.43 & 38.05 & 52.11 & 32.53 \\
                \rowcolor{gray!10} LidaRefer-MP & \checkmark & \textbf{60.14} & \textbf{52.40} & \textbf{59.73} & \textbf{47.89} \\
                \bottomrule
            \end{tabular}%
        }
    \end{center}
    \caption{Performance comparison on Talk2Car-3D. ‘-L’ and ‘-M’ denote our LiDAR-only and multimodal settings, respectively, while ‘-P’ denotes using pre-trained 3D visual encoders.}
    \label{tab1}
\end{table}

\subsection{Main Results} Table~\ref{tab1} presents the results of the comparison models across various configurations on Talk2Car-3D, where the top and bottom parts are for LiDAR-only and multimodal settings, respectively. In the LiDAR-only setting, LidaRefer-L, our base model with a visual encoder trained from scratch, significantly outperforms comparison models in the same setting across all evaluation metrics. Interestingly, it surpasses even the MSSG-LP in terms of broader target identification ability (indicated by ACC@0.25). This demonstrates the effectiveness of our transformer-based architecture and the proposed DiSCo supervision in enhancing both cross-modal alignment and contextual understanding. Furthermore, when employing a pre-trained visual encoder (LidaRefer-LP), the observed performance gain is relatively modest compared to the improvement observed in MSSG-LP over its base model. This suggests that our base architecture, potentially benefiting from the diverse object information learned via the OFS and the contextual relationships modeled by DiSCo, effectively captures a substantial understanding of the visual scene even without pre-trained weights. In the multimodal setting, we achieve similar results, without exception, by incorporating image texture information. The effectiveness stems from RGB images providing complementary features to distinguish visually similar objects and understand complex spatial configurations. 

\subsection{Analysis}

\paragraph{Effects of OFS and DiSCo.} Table~\ref{tab2} presents the ablation results to investigate the impact of OFS and DiSCo. To exploit all BEV features as visual tokens without OFS, we double the voxel size along the X and Y axes to reduce the resolution of the BEV feature map to a level manageable within GPU memory\footnote{Four NVIDIA A6000 GPUs, each with 48GB of memory, are used.}. However, this coarse resolution degrades the ability to detect small objects, leading to a significant performance drop. Notably, the absence of both DiSCo and OFS results in the most substantial performance degradation across all configurations, highlighting their complementary importance. The results show that DiSCo alone yields better performance than OFS alone. Thus, DiSCo has a more substantial impact than OFS.

\begin{table}[t]
    \begin{center}
        \scriptsize{%
            \begin{tabular}{lccccc}
    \toprule
    \multirow{2}{*}{Config} & \multirow{2}{*}{Voxel size} & \multirow{2}{*}{OFS} & \multirow{2}{*}{DiSCo} & \multicolumn{2}{c}{3D} \\
    \cmidrule(lr){5-6}
    & & & & ACC@0.25 & ACC@0.5 \\
    \midrule
    \rowcolor{gray!10} LidaRefer-L & $(0.075, 0.075, 0.2)m$ & \checkmark & \checkmark & \textbf{50.90} & \textbf{31.43} \\
    \midrule 
    \rowcolor{gray!10} LidaRefer-L & $(0.15, 0.15, 0.2)m$ & \checkmark & \checkmark & 48.39 & 28.51 \\
    w/o OFS & $(0.15, 0.15, 0.2)m$ & & \checkmark & 45.98 & 25.90 \\
    w/o DiSCo & $(0.15, 0.15, 0.2)m$ & \checkmark & & 45.48 & 24.20 \\
    w/o both & $(0.15, 0.15, 0.2)m$ & & & 40.06 & 21.69 \\
    \bottomrule
\end{tabular}% 
        }
    \end{center}
    \caption{The ablation study of our proposed modules}
    \label{tab2}
\end{table}

\begin{table}[t]
    \begin{center}
        \scriptsize{%
            \begin{tabular}{lcccc}
                \toprule
                \multirow{2}{*}{Method} & \multirow{2}{*}{OFS} & \multirow{2}{*}{DiSCo} & \multicolumn{2}{c}{3D} \\
                \cmidrule(lr){4-5} % 3D (4-5열) 아래에 선 추가
                & & & ACC@0.25 & ACC@0.5 \\
                \midrule
                MSSG \cite{Cheng2023} & & & 41.97 & 24.10 \\
                + OFS & \checkmark & & 43.07 & 27.91 \\
                + DiSCo & & \checkmark & 43.27 & 27.10 \\
                + both & \checkmark & \checkmark & 44.38 & 29.91 \\
                \midrule % 그룹 구분을 위해 추가
                \rowcolor{gray!10} LidaRefer-L (Ours) & \checkmark & \checkmark & \textbf{50.90} & \textbf{31.43} \\
                \bottomrule
            \end{tabular}% 
        }
    \end{center}
    \caption{Extension of both OFS and DiSCo to existing outdoor 3D VG models}
    \label{tab3}
\end{table}

\paragraph{Plugging into the existing method.} Our OFS and DiSCo are designed to be plug-and-play modules that can be seamlessly integrated into the existing 3D VG models without requiring major architectural changes. To evaluate their general applicability, we incorporate both modules into MSSG and report the results in Table~\ref{tab3}. Adding either OFS or DiSCo individually leads to consistent performance gains over the original MSSG, and their combination yields further improvements, demonstrating effectiveness and flexibility. However, the MSSG variants underperform compared to LidaRefer-L. This performance gap is primarily attributed to architectural limitations inherent in MSSG. Specifically, MSSG's coarse-grained fusion approach relies solely on sentence-level features, potentially losing specific word meanings in complex descriptions, while its architecture insufficiently addresses contextual understanding needed to fully utilize DiSCo's spatial relationship modeling.

\section{Conclusion}
We propose LidaRefer, a context-aware 3D VG framework tailored for large-scale outdoor scenes. It addresses two challenges, background-dominated LiDAR inputs and lack of non-target annotations, by extracting object-centric features and modeling referential context via DiSCo, a supervision method that captures spatial relationships without manual labeling. Our approach achieves state-of-the-art results on Talk2Car-3D, validating its effectiveness in large-scale outdoor scenarios.

\newpage

\appendix

\section*{Appendix}
\label{sec:supplementary_material_title}

This supplementary material provides comprehensive information to support our paper. Section~\ref{appendix_disco} explains the DiSCo algorithm and pseudo-labeling strategy. Section~\ref{appendix_implementation} elaborates on the implementation details, including dataset construction, training procedures, and comparison models. Finally, Section~\ref{appendix_analysis} presents further quantitative and qualitative analyses.

\section{DiSCo with Pseudo-labeling strategy}
\label{appendix_disco}
As Algorithm~\ref{alg:disco_corrected}, DiSCo explicitly models spatial relationships to enhance referential context understanding with the pseudo-labeling strategy.

\begin{algorithm}[ht]
\centering
\small  
\caption{DiSCo: Discriminative-Supportive Collaborative localization}
\label{alg:disco_corrected} 

\begin{tabular}{p{0.967\textwidth}} 
\hline
\textbf{input:} \\
$\bullet$ $Q = \{q_i\}_{i=1}^{K}$: decoder queries. \\
$\bullet$ $B = \{b_j\}_{j=1}^{N}$: 3D bounding boxes for all object proposals in a scene. \\
$\bullet$ $b^t$: 3D bounding box for target object. \\
$\bullet$ $\tau$: distance threshold for matching. \\
\hline
\textbf{functions:} \\
$\bullet$ $\text{Center}(x)$: extracts the center coordinate. \\
$\bullet$ $\text{Distance}(p_1, p_2)$: calculates Euclidean distance between two points. \\
$\bullet$ $\text{Localize}(X)$: performs localization for the set $X$. \\
\hline
\end{tabular}

\begin{algorithmic}[1] 
\Function{DiSCo}{$B, Q, b^t, \tau$}
    \State $R \leftarrow \emptyset$ \Comment{Initialize empty set of referential non-target objects}
    \For{$k = 1$ to $K$}
        \State $b^* \leftarrow \text{argmin}_{b_j \in B} \text{Distance}(\text{Center}(q_k), \text{Center}(b_j))$ \Comment{Find closest object} 
        \State $d \leftarrow \text{Distance}(\text{Center}(q_k), \text{Center}(b^*))$
        \If{$d < \tau$}
            \State $R \leftarrow R \cup \{b^*\}$ \Comment{Add to referential non-target objects if within threshold}
        \EndIf
    \EndFor
    \State \Return Localize($\{b^t\} \cup R$) \Comment{Jointly localize both target and referential non-target objects}
\EndFunction
\end{algorithmic}
\end{algorithm}

\section{Implementation Details}
\label{appendix_implementation}
\paragraph{Dataset.} Based on Talk2Car \cite{Deruyttere2019}, originally introduced as a 2D visual grounding dataset, we construct Talk2car-3D following MSSG's \cite{Cheng2023} preprocessing approach to extend its applicability to 3D scenarios. We refine the target objects from Talk2Car for our training and inference based on three criteria: (1) inclusion in the 10 super categories covered in the nuScenes object detection task, (2) presence within the detection range specific to each category in nuScenes, and (3) containment of at least one point in their ground-truth 3D bounding boxes. For each scenario, we utilize visual data consisting of point clouds from 10 sweeps and their corresponding image, following nuScenes protocol. 

To train LidaRefer, we require additional annotations that are not provided in original Talk2Car dataset. These include (1) categories and center coordinates of all non-target objects for object-centric feature selection (OFS), and (2) 3D bounding boxes of all non-target objects used in DiSCo. These annotations are retrieved from 3D object detection labels in nuScenes. We retain only those non-target objects whose 3D bounding box centers project onto the front camera images since Talk2Car focuses on the front camera perspective.

\paragraph{Training.} We employ the same attention mechanism as \cite{Jain2022, Wu2023} within the cross-modality encoder and decoder, which consists of $N_E=1$ and $N_D=3$ layers, respectively. The number of visual tokens (i.e., object-centric features) and decoder queries is 500 and 256, respectively. The detection range is set to $[-54,54]m$, $[0, 54]m$, and $[-5, 3]m$ along the X, Y, and Z axes, with a voxel size of $(0.075m, 0.075m, 0.2m)$. To improve model robustness against real-world variations, we apply global scaling and translation augmentations to both point clouds and RGB images. All models are trained for 20 epochs with a batch size of 16 on 4 NVIDIA A6000 GPUs, using Adam \cite{Kingma2015} optimizer with a one-cycle policy, maximum learning rate of 4e-4, weight decay of 0.01, and momentum from 0.85 to 0.95.

LidaRefer is trained using a weighted sum of four different loss functions. First, in OFS, a general heatmap classification loss $\mathcal{L}_{hm}$ is used to supervise the center heatmap head to predict the center positions of all objects in a scene. Second, a query proposal loss $\mathcal{L}_{qp}$ is used to supervise the query proposal network to predict a target query from text-aligned visual features. Third, a target classification loss $\mathcal{L}_{cls}$ is used to supervise the target identification head in the decoder to determine which refined query corresponds to the target. Lastly, a box regression loss $\mathcal{L}_{reg}$ is used to supervise the box regression head to predict the box's center, size, and rotation for both the target and referential objects. We adopt focal loss \cite{Lin2020} for heatmap, query proposal, and target classification, and L1 loss for box regression. The final loss is defined as: \begin{equation} \mathcal{L} = \lambda_1 \mathcal{L}_{hm} + \lambda_2 \mathcal{L}_{qp} + \lambda_3 \mathcal{L}_{cls} + \lambda_4 \mathcal{L}_{reg}
\label{eq:final_loss} \end{equation} where the weights of four losses $\lambda_1, \lambda_2, \lambda_3, \lambda_4$ are set to 1, 0.5, 0.5, 1.25, respectively.

For training our center heatmap head, query proposal network and target identification network, we employ a center-based object detector's assignment strategy \cite{Zhou2022, Yin2021}. This strategy constrains each network to learn information exclusively about objects whose centers align with the positions of the selected features. Thus, to achieve stable and faster training convergence, we manually select relevant features necessary for subsequent modules during the OFS and query initialization stages. Specifically, in the OFS, we first ensure selected object-centric visual features $\mathcal{F}_{\mathcal{V}}$ retain information about all objects within a scene by manually incorporating the features from $\mathcal{F}_{\mathrm{BEV}}$ that are positioned at the central locations of all objects into $\mathcal{F}_{\mathcal{V}}$. Then, the remaining features of $\mathcal{F}_{\mathcal{V}}$ are chosen based on the heatmap scores predicted by the model from $\mathcal{F}_{\mathrm{BEV}}$. Likewise, in the query initialization, we ensure decoder queries $Q$ retain information about the target object by manually incorporating the feature positioned at the central location of the target object.

\paragraph{Comparison Models.} We compare LidaRefer against recent state-of-the-art outdoor 3D visual grounding baselines: MSSG and BEVGrounding. 
\begin{itemize}
    \item \textbf{MSSG} \cite{Cheng2023}: is a pioneering outdoor 3D VG model for autonomous driving. It first generates a BEV feature map from point clouds and fuses it with sentence-level text features, extracted by a 2-layer BERT \cite{Devlin2019}, using window-based self-attention from Swin Transformer \cite{liu2021swin}. It then employs a standard center-based detection head to identify the target object, selecting the box generated from the highest-scoring position.  
    \item \textbf{BEVGrounding} \cite{liu2024talk}: includes two variants: two-stage and one-stage. The two-stage variant first generates object proposals using pre-trained BEVFusion \cite{liu2023bevfusion} and extracts sentence-level text features using CLIP \cite{radford2021learning} text encoder. Then, it identifies the target object through a lightweight matching network. The one-stage variant employs an end-to-end transformer-based architecture with a trimodal BEV encoder for global feature fusion and a grounding decoder using attention mechanisms to predict the target object directly.
\end{itemize}

\section{Additional Analysis}
\label{appendix_analysis}
\begin{table}[h] 
    \begin{minipage}[t]{0.48\textwidth} 
        \begin{center}
        \scriptsize{% 
            \begin{tabular}{lcc}
    \toprule
    \multirow{2}{*}{Visual Token Num} & \multicolumn{2}{c}{3D} \\
    \cmidrule(lr){2-3} % 3D (2-3열) 아래에 선 추가
    & ACC@0.25 & ACC@0.5 \\
    \midrule
    300 & 49.00 & 27.31 \\
    400 & 50.60 & 29.41 \\
    \rowcolor{gray!10} 500 & \textbf{50.90} & \textbf{31.43} \\
    600 & 47.49 & 29.82 \\
    700 & 46.39 & 29.12 \\
    \bottomrule
\end{tabular}% 
        }
        \end{center}
        \caption{Performance comparison of different visual token numbers}
        \label{tab:supp_visual_tokens}
    \end{minipage}%   
    \hfill         
    \begin{minipage}[t]{0.48\textwidth} 
        \begin{center}
        \scriptsize{%
            \begin{tabular}{lcc}
    \toprule
    \multirow{2}{*}{Decoder Query Num} & \multicolumn{2}{c}{3D} \\
    \cmidrule(lr){2-3} % 3D (2-3열) 아래에 선 추가
    & ACC@0.25 & ACC@0.5 \\
    \midrule
    64 & 49.00 & 27.31 \\
    128 & 50.20 & 31.12 \\
    \rowcolor{gray!10} 256 & \textbf{50.90} & \textbf{31.43} \\
    384 & 48.49 & 28.31 \\
    \bottomrule
\end{tabular}% 
        }
        \end{center}
        \caption{Performance comparison of different decoder query numbers}
        \label{tab:supp_decoder_queries}
    \end{minipage}
\end{table}

\paragraph{Sizes of Visual Tokens and Decoder Queries.} Tables~\ref{tab:supp_visual_tokens} and~\ref{tab:supp_decoder_queries} show the impact of varying the number of visual tokens and decoder queries, respectively. The results indicate that performance improves only when these quantities fall within an optimal range. Deviating from this range, either by increasing or decreasing the number excessively, leads to degradation. For visual tokens, using too few limits the ability to capture global context, while too many introduce unnecessary background noise, hindering cross-modal alignment and contextual understanding. Similarly, decoder queries must be carefully balanced. An insufficient number limits the capacity to learn diverse referential contexts, whereas an excessive number disperses attention across irrelevant proposals, weakening the ability to disambiguate the target object.

\begin{figure}[t]
  \centering
  \includegraphics[width = 0.95\textwidth]{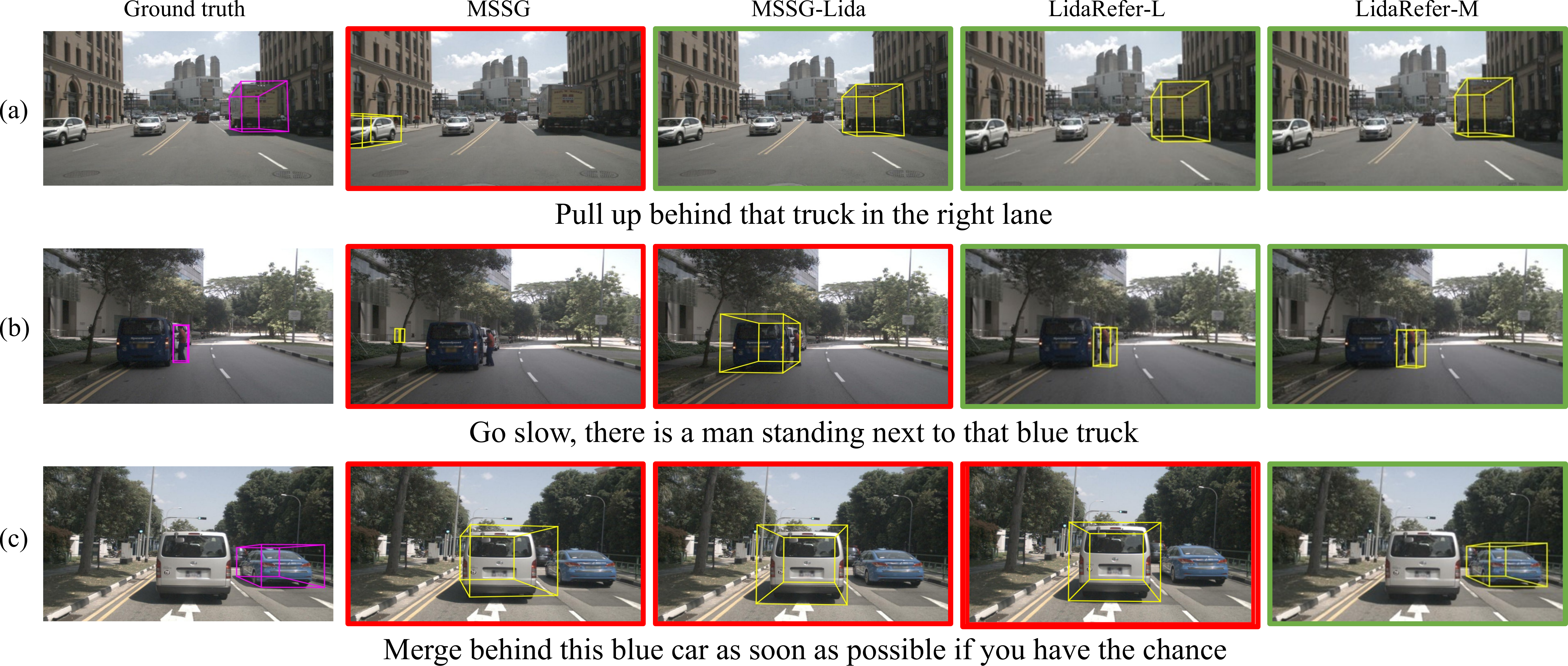} 
  \caption{Visualization comparison between MSSG-based and LidaRefer-based methods for various scenarios. The ground-truth and prediction boxes within the scenes are in pink and yellow, respectively. MSSG-Lida refers to MSSG enhanced with OFS and DiSCo from LidaRefer. Under an IoU threshold setting of 0.5, the scenes in a green outline are correct predictions, whereas red outlines indicate incorrect ones.}
  \label{fig:supp_qualitative_comparison}
\end{figure}

\paragraph{Qualitative Results.}
Fig.~\ref{fig:supp_qualitative_comparison} illustrates the qualitative comparisons between MSSG variants and LidaRefer, highlighting the advantages of LidaRefer in various challenging scenarios. Fig.~\ref{fig:supp_qualitative_comparison}(a) shows that incorporating OFS and DiSCo, which leverage non-target object information, enhances the model’s ability to distinguish object categories and accurately identify the target. Fig.~\ref{fig:supp_qualitative_comparison}(b) demonstrates that LidaRefer-based methods effectively understand contextual information in challenging scenarios where multiple objects within the same category as the target appear in a visual scene and a language description mentions multiple objects. In contrast, MSSG-based methods incorrectly identify non-target objects that either share a category with the target or are mentioned in the description. Fig.~\ref{fig:supp_qualitative_comparison}(c) shows that LidaRefer-M effectively utilizes RGB information for accurate target identification in scenarios where distinguishing the target based on color is essential.

\bibliography{egbib}
\end{document}